\documentclass[a4paper]{article}
\usepackage{graphicx}
\usepackage{twocolceurws}
\usepackage{caption}
\usepackage{subcaption}
\usepackage{algorithm}
\usepackage{algorithmic}
\usepackage{multicol}
\usepackage{booktabs}
\usepackage{multirow}
\usepackage{makecell}
\usepackage{graphicx}
\usepackage[table]{xcolor}
\usepackage{flushend}
\usepackage{amsmath}
\usepackage{makecell}

\usepackage{blindtext}
\usepackage{enumitem}
\usepackage{amsmath}
\usepackage[affil-it]{authblk}
\makeatletter
\def\thanks#1{\protected@xdef\@thanks{\@thanks
        \protect\footnotetext{#1}}}
\makeatother

\title{DPAttack: Diffused Patch Attacks against Universal Object Detection}
\author[1]{Shudeng Wu}
\author[1,2,*]{Tao Dai}
\author[1,2,*]{Shu-Tao Xia}

\affil[1]{Tsinghua Shenzhen International Graduate School, Tsinghua University, Shenzhen, China }
\affil[2]{PCL Research Center of Networks and Communications, \protect\\ Peng Cheng Laboratory, Shenzhen, China}
\affil[ ]{wsd19@mails.tsinghua.edu.cn, daitao.edu@gmail.com, xiast@sz.tsinghua.edu.cn \thanks{*Corresponding author: Tao Dai and Shu-Tao Xia}}
\institution{}
\begin{document}
\maketitle
\begin{abstract}
     Recently, deep neural networks (DNNs) have been widely and successfully used in Object Detection, e.g. Faster RCNN, YOLO, CenterNet. However, recent studies have shown that DNNs are vulnerable to adversarial attacks. Adversarial attacks against object detection can be divided into two categories, whole-pixel attacks and patch attacks. While these attacks add perturbations to a large number of pixels in images, we proposed a diffused patch attack (\textbf{DPAttack}) to successfully fool object detectors by diffused patches of asteroid-shaped or grid-shape, which only change a small number of pixels. Experiments show that our DPAttack can successfully fool most object detectors with diffused patches and we get the second place in the Alibaba Tianchi competition:  Alibaba-Tsinghua Adversarial Challenge on Object Detection. Our code can be obtained from https://github.com/Wu-Shudeng/DPAttack.
\end{abstract}


\section{Introduction}
Object detection aims to locate objects (e.g. persons, dogs, flowers) from images. Recently deep neural networks (DNNs) \cite{ren2015faster, cai2018cascade, duan2019centernet, liu2016ssd, bochkovskiy2020yolov4, lin2017focal} have been widely and successfully used in object detection, which can be categorized into  two-stage and one-stage methods. Faster RCNN \cite{ren2015faster} and Cascade RCNN \cite{cai2018cascade} are two-stage methods that first use region proposal network (RPN) to obtain thousands of proposals and then classify these proposals into different classes. YOLO \cite{bochkovskiy2020yolov4} and SSD \cite{liu2016ssd} are one-stage methods which directly regress object bounding boxes and classify them.

One critical difference between two-stage and one-stage methods is that the sizes of their feature map are quite different. The feature map of Faster RCNN is down-sampled by $4\times$ from input images while that of YOLOv4 is down-sampled by $32\times$. As a consequence, the features of Faster RCNN have quite a smaller receptive field than those of YOLOv4. Meanwhile, two-stage methods like Faster RCNN contain an RoI pooling operation, which attends to all features within a proposal region. These features have a smaller receptive field and have equal contributions for proposal classification, so as to show more robustness to local perturbation. As our experiments have shown, two-stage detectors are harder to be attacked than one-stage detectors.

\begin{figure}[t]
    \begin{center}$
    \begin{array}{ccccc}
    &
    \includegraphics[width=0.15\textwidth]{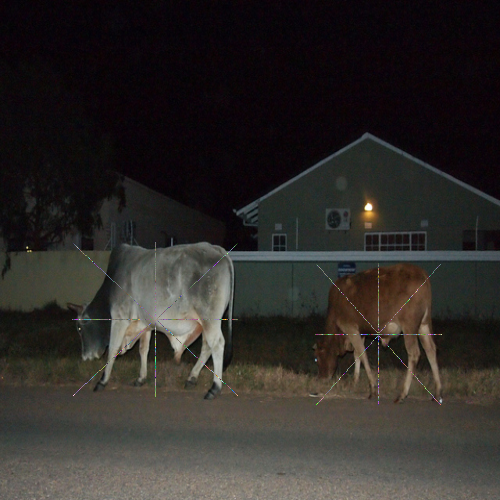} &
    \includegraphics[width=0.15\textwidth]{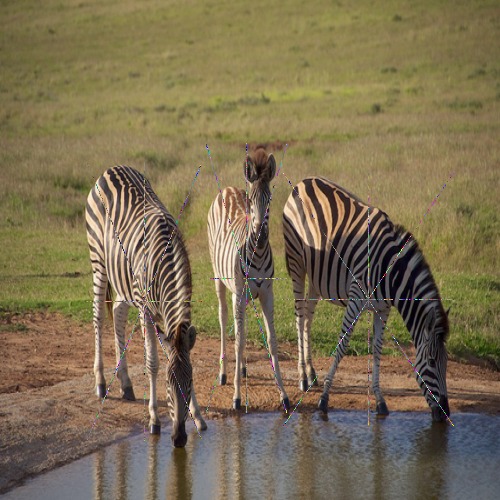}& \\
    &
    \includegraphics[width=0.15\textwidth]{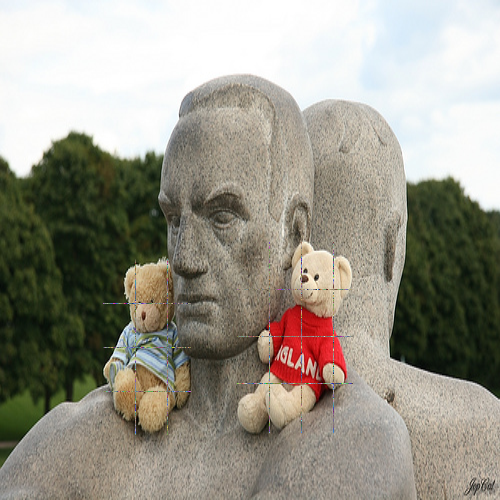}&
    \includegraphics[width=0.15\textwidth]{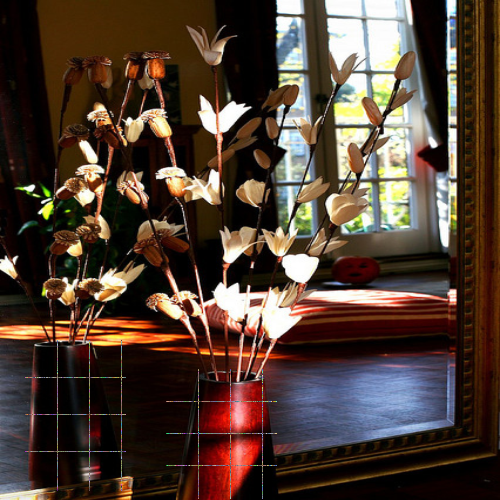} \\
    \end{array}$
    \end{center}
    \caption{Adversarial patches of asteroid-shaped (first row) and grid-shape (second row).}
    \label{fig:patch_examples}
\end{figure}

Adversarial attacks can fool models in lots of computer vision tasks, e.g. images classification \cite{goodfellow2014explaining, carlini2017towards, moosavi2016deepfool}, images segmentation \cite{li2018robust, hendrik2017universal}, face detection \cite{yang2019design, bose2018adversarial} and object detection \cite{xie2017adversarial, wei2018transferable, lee2019physical, liu2018dpatch}. Adversarial attacks against object detection can be divided into two categories: 1) whole-pixel attacks, which can add perturbations to all pixels in images under $L_p$ constraints (e.g. $L_1, L_{\infty}$); 2) patch attacks, which add perturbations to local pixels of a quite small area in images. However, Both whole-pixel attacks and patch attacks require a large area of images to add perturbations, which can be easily detected by human eyes or detectors. So we aim to change as fewer pixels as possible and make the number of patches (connected domains of changed pixels) less than 10. 


To make our perturbation affect more features and at the same time change as fewer pixels as possible, we design a diffused patch of asteroid-shaped or grid-shaped (as shown in \textbf{Figure \ref{fig:patch_examples}}). Specifically, we create asteroid-shaped or grid-shaped patches inside the bounding boxes of objects and using gradient-based methods (i.e. FGSM \cite{goodfellow2014explaining}) to update these patches iteratively. Meanwhile, both two-stage and one-stage detectors have nearly thousands of proposals. To avoid any proposal from being not attacked, we designed a special loss that pays more attention to unsuccessfully attacked proposals and can suppress introducing false positive proposals.
We conclude the main contributions of this paper as follow:

\begin{itemize}
    \item We design diffused patches of asteroid-shaped or grid-shaped, which can affect more features in the feature map of detectors at the cost of only changing a small number of pixels.
    \item Our attacking loss pays more attention to unsuccessfully attacked proposals and can suppress introducing false positive proposals.
    \item Experiments show that our method can successfully fool both two-stage and one-stage detectors at the cost of changing a small number of pixels. We get the second place in the Alibaba Tianchi competition:  Alibaba-Tsinghua Adversarial Challenge on Object Detection.
\end{itemize}

\section{Methodology}

\subsection{Problem Formulation}
Two-stage object detectors use region proposal network (RPN) to obtain thousands of proposals and then classify them. One-stage detectors directly regress bounding boxes and scores of objects simultaneously and we define the pair of a bounding box and a score as a proposal as well. To this end, we define $P=\{p_1, p_2, ..., p_N\}$ as proposals for both two-stage and one-stage detectors. Each element $P_i=\{x, y, h, w, s\}$ in $P$ contains 5 components: left-top position, height, width of a bounding box, and scores of all predefined classes. Our attack only uses the scores of predefined classes and we use $S=\{s_i, ..., s_N | s_i \in R^C\}$ to be the corresponding scores for proposals in $P$ (C is the number of classes).

In order to affect more features in the feature map, we attach diffused patched to images. Besides, The patches must satisfy two conditions: 1) the number of patches is less than 10; 2) the total number of changed pixels is less than 2\% of the total number of pixels in an image.

\subsection{DPAttack: Diffused Patch Attack}
In this section, we introduce our proposed diffused patch attack in detail. Choosing the positions of patches is the first problem of the diffused patch attack. Intuitively, attaching the patches inside the bounding boxes is most effective to fool detectors and our visualization of gradients verifies it. To this end, we use the centers of bounding boxes as centers of patch masks of asteroid-shaped or grid-shaped (as shown in \textbf{Figure \ref{fig:patch_examples})}. The attacking loss aims to make scores of classes below a threshold and at the same time suppress introducing false positive proposals. We formulate the attacking loss as 
\begin{equation}
    L(x , M, \delta) = \sum_{i=1}^{N} \sum_{c=1}^{C} max (0, f_i^{c}(x\cdot (1-M) + \delta\cdot M)-t)
\end{equation}

Here $x\in R^{3\times h\times w}, M\in R^{3\times h\times w}, \delta\in R^{3\times h\times w}$ means input image, patch masks and values of patches respectively. $f_{i}^{c}(\cdot)$ is the function which generates the score for class $c$ of the $i$-th proposal, and $t$ is the threshold score to distinguish the categories of objects (excluding background). Only positive proposals can contribute to the loss, and false-positive proposals can be effectively suppressed. The details of the diffused patch attack can be referred to as \textbf{Algorithm \ref{alg:1}}.

\begin{algorithm}
    \caption{Diffused Patch Attack}
    \begin{flushleft}
    \hspace*{\algorithmicindent} \textbf{Input:} model function $f(\cdot)$; an input image $x$; bounding boxes $bboxes$; maximum iterations $T$; score threshold $t$; step $\alpha$. \\
    \end{flushleft}
    \begin{flushleft}
    \hspace*{\algorithmicindent} \textbf{Output:} the adversarial example \(x^{'}\)\\
    \end{flushleft}
    
    \begin{algorithmic}[1]
    \STATE generate mask $M$ by taking centers of $bboxes$ as centers of the masks.
    \STATE $\delta \leftarrow \textbf{0}, i \leftarrow 0, n \leftarrow 1$
    \WHILE{$i<T$\ \textbf{and} \ $n > 0$}
    \STATE \(l \leftarrow L(x , M, \delta) 
    \)
    \STATE \(\delta \leftarrow \delta - \alpha \cdot sign(\nabla_{\delta}L(x , M, \delta))\)
    \STATE \(\delta \leftarrow max(0, min(\delta, 255))\)
    \STATE $n \leftarrow$ number of positive proposals
    \STATE \(i \leftarrow i+1\)
    \ENDWHILE
    \STATE $x^{'} \leftarrow x\cdot (1-M) + \delta\cdot M$
    \RETURN \(x^{'}\)
    \end{algorithmic}
    \label{alg:1}
\end{algorithm}

\section{Experiments}

\subsection{Dataset and Evaluation Metrics}
The dataset is provided by the Alibaba Tianchi competition, Alibaba-Tsinghua Adversarial Challenge on Object Detection, and consists of 1000 images ($500\times 500$ in resolution) from test data of MSCOCO 2017. 

We use multiple metrics to evaluate our proposed method. The first metric is overall score (\textbf{OS}) which have two aspects of consideration, number of suppressed bounding boxes and number of changed pixels. 
\begin{equation}
    S(x, x^{'}) = (2-\frac{\sum_{k}R_k}{5000})\cdot (1-\frac{min(BB(x), BB(x^{'}))}{BB(x)})
    \label{eq:1}
\end{equation}

Here $R_k$ is the number of perturbing pixels of $k$-th patch, $BB(\cdot)$ is the number of bounding box of objects in images. Perturbing less number of pixels ($R_k$) and suppressing more bounding boxes of adversarial examples $BB(x^{'})$ can make overall score higher. Besides, we use success rate (\textbf{SR}) to evaluate the performance of our attack. We deem our attack successful if we suppress all bounding boxes in images. The ratio of bounding box (\textbf{BBR}) is the ratio between the number of bounding boxes with regard to original images and adversarial examples. \textbf{APP} is the \textbf{a}verage ratio between number of \textbf{p}erturbing \textbf{p}ixels and whole pixels of adversarial examples. 
\begin{equation}
    \begin{aligned}
        BBR = \frac{\sum_{i=1}^N BB(x_i^{'})}{\sum_{i=1}^N BB(x_i)}\\
        APP = \frac{1}{N} \sum_{i=1}^N \frac{\sum_{k} R_k }{500\times 500}
    \end{aligned}
\end{equation}

\subsection{Overall Results}

\begin{table}[ht]
    \begin{center}
    \caption{Results of diffused patch attack of different shape against YoLov4 \cite{bochkovskiy2020yolov4} and Faster RCNN \cite{ren2015faster}. The number $s$ in Asteroid-$s$ means scaling the size of bounding box of patches by $s$, e.g. 0.8. The numble $l$ in grid-$l\times l$ means that there are $l$ horizontal and vertical lines in the grid-shape patches. Ensemble means that we choose the adversarial examples of the highest overall score (OS) from all these kinds of patches.}
    \label{tab:1}
    \resizebox{\columnwidth}{!}{%
    \begin{tabular}{lcccccccc}
        \toprule
        \multirow{2}{*}[-4pt]{\thead{}} &
        \multicolumn{4}{c}{\thead{YOLOv4\cite{bochkovskiy2020yolov4}}} &
        \multicolumn{4}{c}{\thead{Faster RCNN \cite{ren2015faster}}} \\
        \cmidrule(lr){2-5}
        \cmidrule(lr){6-9}
        &SR & OS  & BBR & APP & SR & OS & BBR & APP \\
        \midrule
         asteroid-0.8 & 96.2 \% & 1515 & 2.76 \% & 0.89 \% & 65.4 \%  & 1241 &12.3 \% &1.05 \% \\
        asteroid-1.0 & 100 \% & 1401 & 0 \% & 1.14 \% & 78.8 \%  & 1203 &7.6 \% &1.27 \% \\
        grid-1x1 & 75 \% & 1423 & 18.62 \% & 0.45 \% & 15.3 \% & 906 &39.5 \% &0.55 \% \\
        grid-2x2 & 100 \% & 1457 & 0 \% & 0.87 \% & 61.5 \% & 1160 &15.38 \% & 1.10 \% \\
        grid-3x3 & 100 \% & 1305 & 0 \% & 1.29 \% & 86.5 \%  & 1076 & 8.6 \% &1.47 \% \\
        grid-4x4 & 100 \% & 1158 & 0 \% & 1.54 \% & 84.6 \% & 1017 &9.38 \% &1.63 \% \\
        ensemble & 98.3 \% & 1563 & 1.82 \% & 0.87 \% &  76.5\% & 1436 &12.1 \% &0.98 \% \\
        \bottomrule
    \end{tabular}%
    }
    \end{center}
\end{table} 

The attack performance of diffused patch attack can be inferred to \textbf{Table \ref{tab:1}}. As our aforementioned analysis, the Faster RCNN \cite{ren2015faster} has features with a small receptive field and its RoI pooling operations take ensemble of features into consideration so as to be more robust than TOLOv4 \cite{bochkovskiy2020yolov4}. The success rates (SR) for YOLOv4 can be nearly 100 \% for most kinds of patches, while less than 85 \% in most cases for Faster RCNN. Similarly, overall scores for YOLOv4 are higher than Faster RCNN, and bounding box ratios (BBR) are lower than Faster RCNN. Besides, the predefined bounding boxes are not the ground truth boxes but are from the results of YOLOv4 and Faster RCNN respectively, so APPs for them are different. In fact, Faster RCNN perturbing more pixels and obtain lower scores.

\subsection{Visualisation of gradients}

Intuitively, it is much more effective to attach the patches inside the bounding boxes of objects in images and we visualize the gradients of attacking loss with regard to the input image. The attacking loss are reformulated from equation (\ref{eq:1}) as follow:
\begin{equation}
    \begin{aligned}
   & p = x\cdot (1-M) + \delta\cdot M \\
   & L(p) = \sum_{i=1}^{N} \sum_{c=1}^{C} max (0, f_i^{c}(p)-t) \\
    \end{aligned}
\end{equation}

\begin{figure}[t]
    \begin{center}$
    \begin{array}{cccc}
    
    \includegraphics[width=0.1\textwidth]{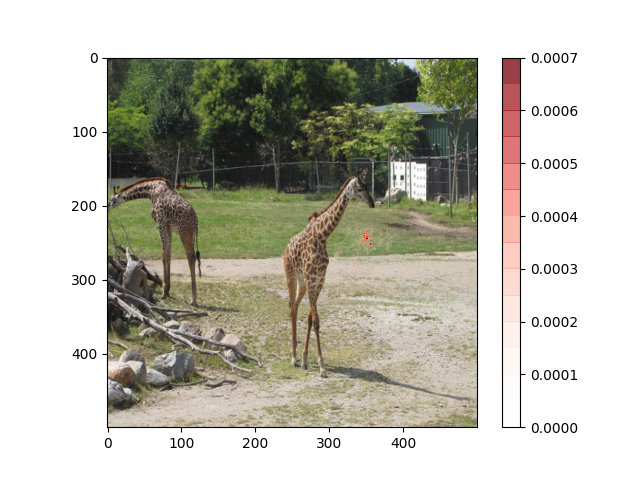}&
    \includegraphics[width=0.1\textwidth]{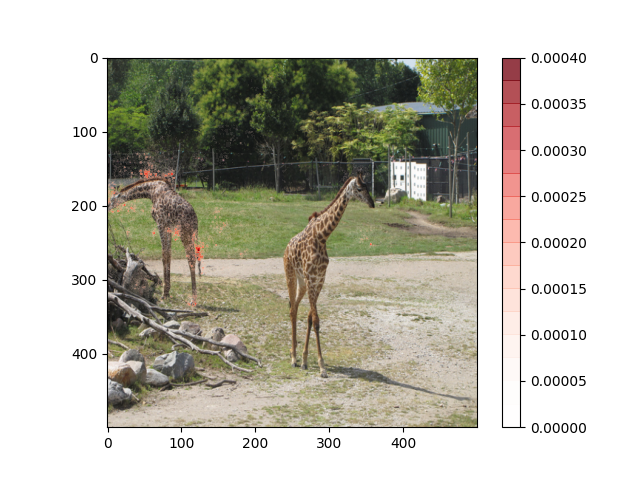}& 
    \includegraphics[width=0.1\textwidth]{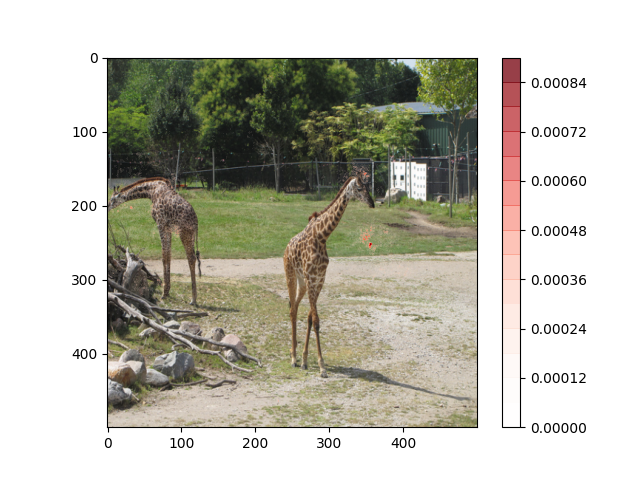}&
    \includegraphics[width=0.1\textwidth]{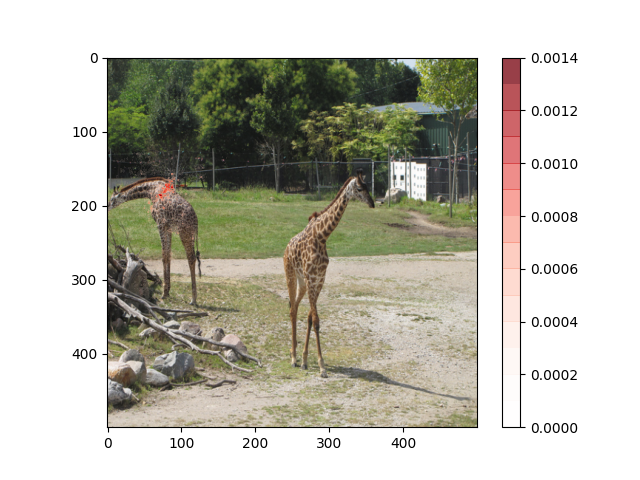}\\
    t=0 & t=30 & t=60 & t=90  \\
    \includegraphics[width=0.1\textwidth]{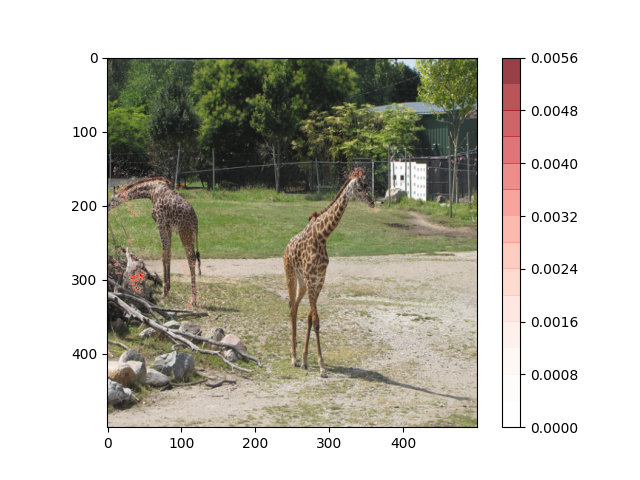} &
    \includegraphics[width=0.1\textwidth]{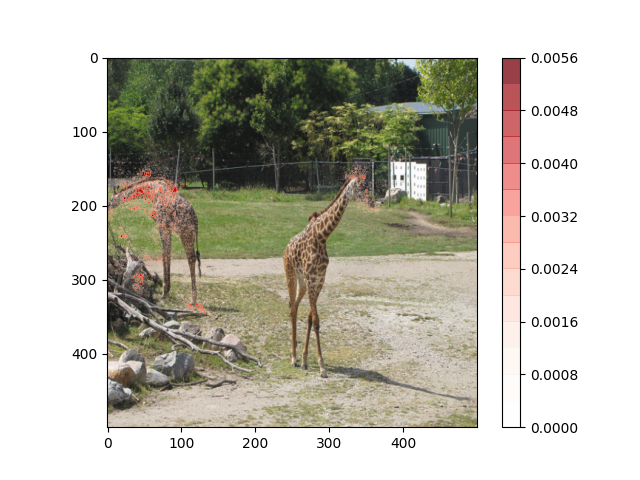}& 
    \includegraphics[width=0.1\textwidth]{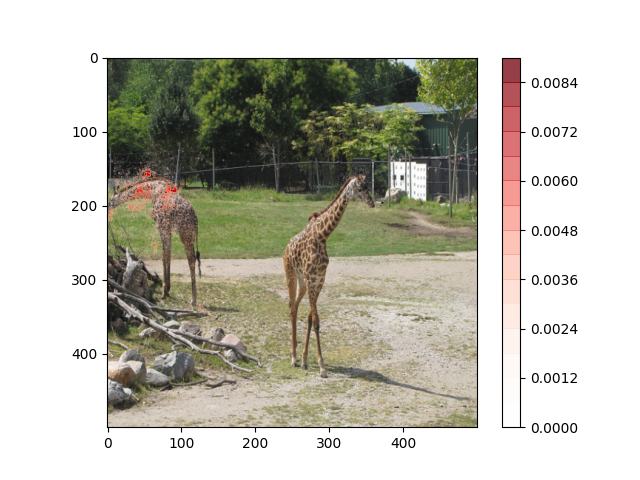}&
    \includegraphics[width=0.1\textwidth]{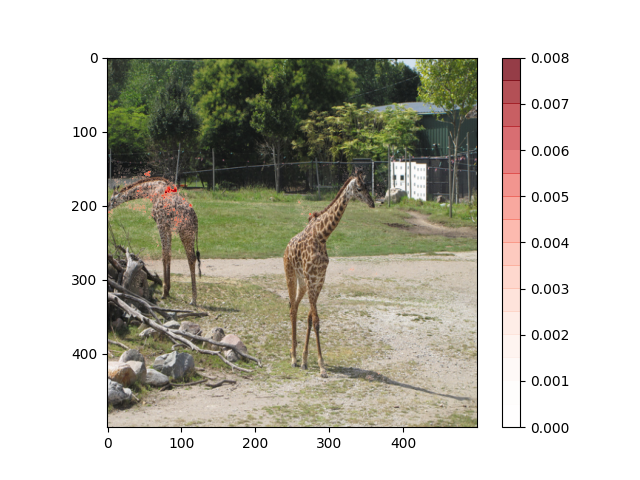}\\
    t=0 & t=40 & t=80 & t=120 \\
    \end{array}$
    \end{center}
    \caption{Visualisation of gradients for YOLOv4 (first row) and Faster RCNN (second row). $t$ means iterations during the attack process.}
    \label{fig:visualisation}
\end{figure}

We visualize the gradients $||\nabla_p L(p)||_1$ by applying  their transparent heatmap to the original image. As show in \textbf{Figure \ref{fig:visualisation}}, the gradients in the bounding boxes of objects are higher than those outside the bounding boxes. So it is more effective to put patches inside the bounding boxes.

\section{Conclusion}
In this paper, we proposed the diffused patch attack of asteroid-shaped or grid-shaped that can successfully fool both one-stage and two-stage detectors. Besides, our attack loss can pay more attention to positive proposals and suppress introducing false positive proposals. Experiments show that our proposed methods can successfully fool detectors at the cost of perturbing a small number of pixels and we get second place in the Alibaba Tianchi competition: Alibaba-Tsinghua Adversarial Challenge on Object Detection.

\bibliographystyle{abbrv}
\bibliography{main}





\end{document}